\documentclass[10pt, a4paper]{article}

\usepackage{lrec-coling2024} 

\usepackage{booktabs}
\usepackage{multirow}
\usepackage{amsmath}
\usepackage{subfigure}

\title{ProtoEM: A Prototype-Enhanced Matching Framework for \\ Event Relation Extraction}

\name{\parbox{\linewidth}{\centering
Zhilei Hu$^{1,2}$, Zixuan Li$^{1,2}$, Daozhu Xu$^{3,4}$, Long Bai$^{1,2}$, Cheng Jin$^{3,4}$, \\ Xiaolong Jin$^{1,2}$, Jiafeng Guo$^{1,2}$, Xueqi Cheng$^{1,2}$}}

\address{
        $^{1}$ CAS Key Laboratory of Network Data Science and Technology, \\
        Institute of Computing Technology, Chinese Academy of Sciences, China;\\
        $^{2}$ School of Computer Science and Technology, University of Chinese Academy of Sciences, China; \\ 
        $^{3}$ State Key Laboratory of Geo-Information Engineering, Xi'an 710054, China; \\
        $^{4}$ Xi'an Research Institute of Surveying and Mapping, Xi'an 710054, China. \\
        \{huzhilei19b, lizixuan, bailong, jinxiaolong, guojiafeng, cxq\}@ict.ac.cn \\}

\abstract{
Event Relation Extraction (ERE) aims to extract multiple kinds of relations among events in texts.
However, existing methods singly categorize event relations as different classes, which are inadequately capturing the intrinsic semantics of these relations.
To comprehensively understand their intrinsic semantics, in this paper, we obtain prototype representations for each type of event relation and propose a Prototype-Enhanced Matching (ProtoEM) framework for the joint extraction of multiple kinds of event relations.
Specifically, ProtoEM extracts event relations in a two-step manner, i.e., prototype representing and prototype matching. 
In the first step, to capture the connotations of different event relations, ProtoEM utilizes examples to represent the prototypes corresponding to these relations.
Subsequently, to capture the interdependence among event relations, it constructs a dependency graph for the prototypes corresponding to these relations and utilized a Graph Neural Network (GNN)-based module for modeling.
In the second step, it obtains the representations of new event pairs and calculates their similarity with those prototypes obtained in the first step to evaluate which types of event relations they belong to.
Experimental results on the MAVEN-ERE dataset demonstrate that the proposed ProtoEM framework can effectively represent the prototypes of event relations and further obtain a significant improvement over baseline models.
 \\ \newline \Keywords{Event Relation Extraction, Prototype Representing, Prototype Matching} 
}

\begin{document}

\maketitleabstract

\section{Introduction}

Event Relation Extraction (ERE) is a task that seeks to predict multiple kinds of relations among events in texts, including temporal, causal, subevent, and coreference relations.
ERE is an important way to construct wide connections among events, which is beneficial to various practical applications, such as event prediction~\citep{Chaturvedi2017StoryComprehensionPredicting, Bai2021IntegratingDeepEventLevel}, reading comprehension~\citep{Berant2014ModelingBiologicalProcesses}, and question answering~\citep{Oh2017MultiColumnConvolutionalNeural, Khashabi2018QuestionAnsweringGlobal}.

\begin{figure*}[ht]
  \centering
  \includegraphics[width=0.9\textwidth]{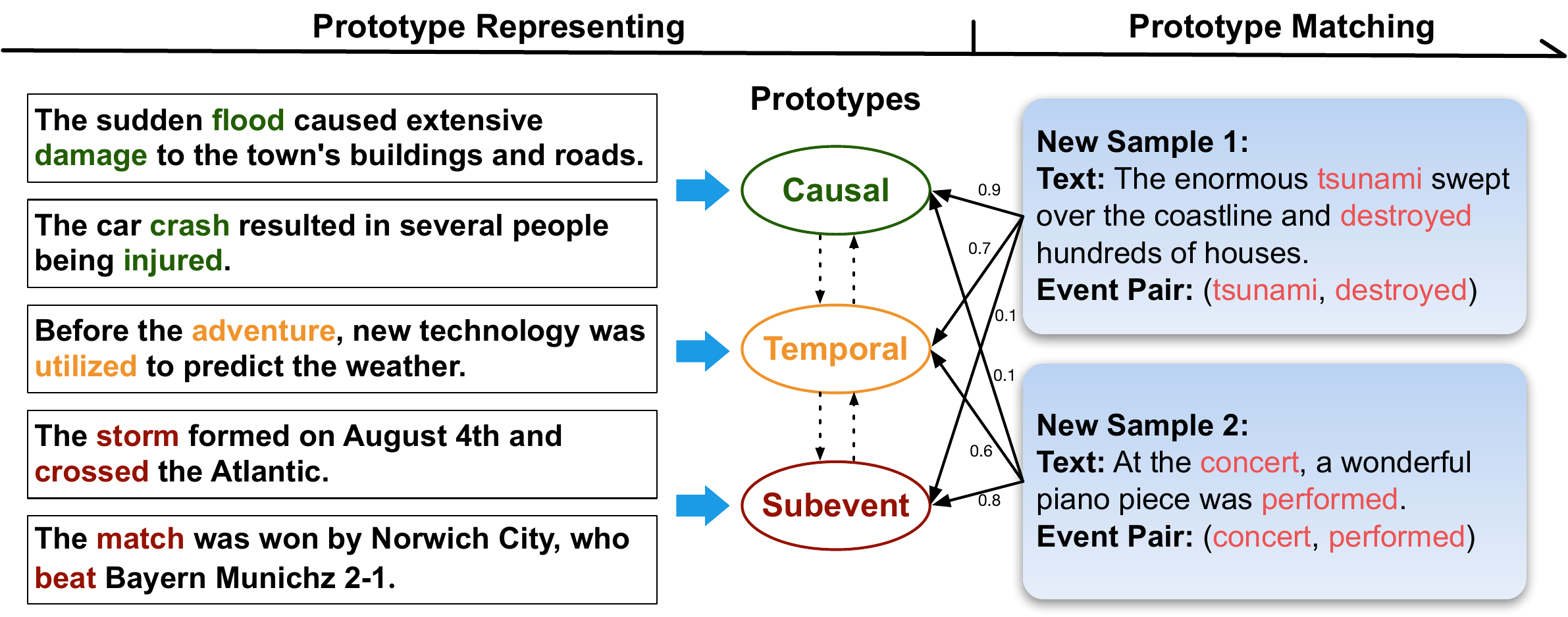}
  \caption{An illustration of the process of prototype representing and prototype matching.}
  \label{fig:intro}
\end{figure*}

ERE is challenging because it requires models to adequately understand the semantics of events and event relations.
Existing works primarily focus on the former, i.e., the semantics of events.
For example, some methods~\citep{Gao2019ModelingDocumentlevelCausal, Phu2021GraphConvolutionalNetworks} build the event graph based on linguistic tools and apply GNNs to enhance the representation of events with their neighbors for final prediction. 
Other methods~\citep{Liu2020KnowledgeEnhancedEvent, Cao2021KnowledgeEnrichedEventCausality} enrich the associations between events by introducing external knowledge, such as events in ConceptNet~\citep{Speer2017ConceptnetOpenMultilingual} that are related to focused events.
In addition, \citet{Chen2022ERGOEventRelational} takes into account the transitivity of event relations by regarding event pairs as nodes and the possible transitivity among event pairs as edges.
However, existing methods singly categorize event relations as different classes, which are inadequately capturing the intrinsic semantics of these relations.

When humans undertake the ERE task, they first comprehend the meaning of event relations from examples.
Inspired by this, we attempt to obtain prototype representations for different types of event relations and explore two kinds of semantic information that are crucial for the comprehensive understanding of event relations, namely, the connotations of prototypes and the interdependence among prototypes.

The first kind of information can be acquired through the examples of each type of event relation, i.e., the examples of its typical applications.
In the examples, two parts of information contribute to event relation extraction: one is the information about the events themselves, and the other is event-agnostic contextual patterns.
For instance, in the case of the event relation ``\textit{cause}'', the typical manner of its application is as follows: ``\textit{The sudden flood caused extensive damage to the town's buildings and roads.}'', there is a ``\textit{cause}'' relation between events ``\textit{flood}'' and ``\textit{damage}''.
Event relations can be reflected not only by event pairs in the examples but also by event-agnostic contexts.
These two categories of connotations contribute to a better representation of prototypes.

For the second kind of information, it can be accessed through structural dependencies among prototypes.
It is most intuitive that causal and temporal relations have strong dependencies.
For example, if a causal relation ``\textit{Cause(flood, damage)}'' exists, it is highly likely that a temporal relation ``\textit{Before(flood, damage)}'' also exists.
Through the interdependence among prototypes, a deeper comprehension of the meaning and characteristics of event relations can be achieved.

After comprehending the meaning of event relations, humans then assess whether a new sample belongs to a given relation by evaluating their similarity.
Likewise, we start by comprehending the contextual meaning of an event pair, then compare it with the prototypes of each event relation to identify the prototype that best aligns with its meaning, thus deriving the relation between the events.
The complete ERE process is illustrated in Figure~\ref{fig:intro}.

Similar to the aforementioned process, we propose a Prototype-Enhanced Matching (ProtoEM) framework for jointly extracting multiple types of event relations.
ProtoEM extracts event relations in a two-step manner, i.e., prototype representing and prototype matching.
In the first step, ProtoEM utilizes the connotation component to model examples corresponding to each type of event relation, thus obtaining prototype representations for each event relation.
It then constructs a dependency graph of different types of prototypes and employs a GNN-based interdependence component to capture their interdependence.
In the second step, it acquires the representations of new event pairs and calculates their similarity with those of prototypes obtained in the first step to evaluate which types of event relations they belong to.

In general, the main contributions of this paper can be summarized as follows:
\begin{itemize}
\item 
We obtain prototype representations for different types of event relations by exploring two kinds of semantic information that are crucial for the comprehensive understanding of event relations, namely, the connotations of prototypes and the interdependence among prototypes.

\item 
We propose a Prototype-Enhanced Matching (ProtoEM) framework for jointly extracting multiple kinds of event relations through a matching process between prototypes and event pairs.
\item 
According to experimental results on the widely used dataset MAVEN-ERE, ProtoEM achieves significant performance improvements across multiple types of event relations compared to the state-of-the-art baselines.
\end{itemize}

\begin{figure*}[t]
  \centering
  \includegraphics[width=0.8\textwidth]{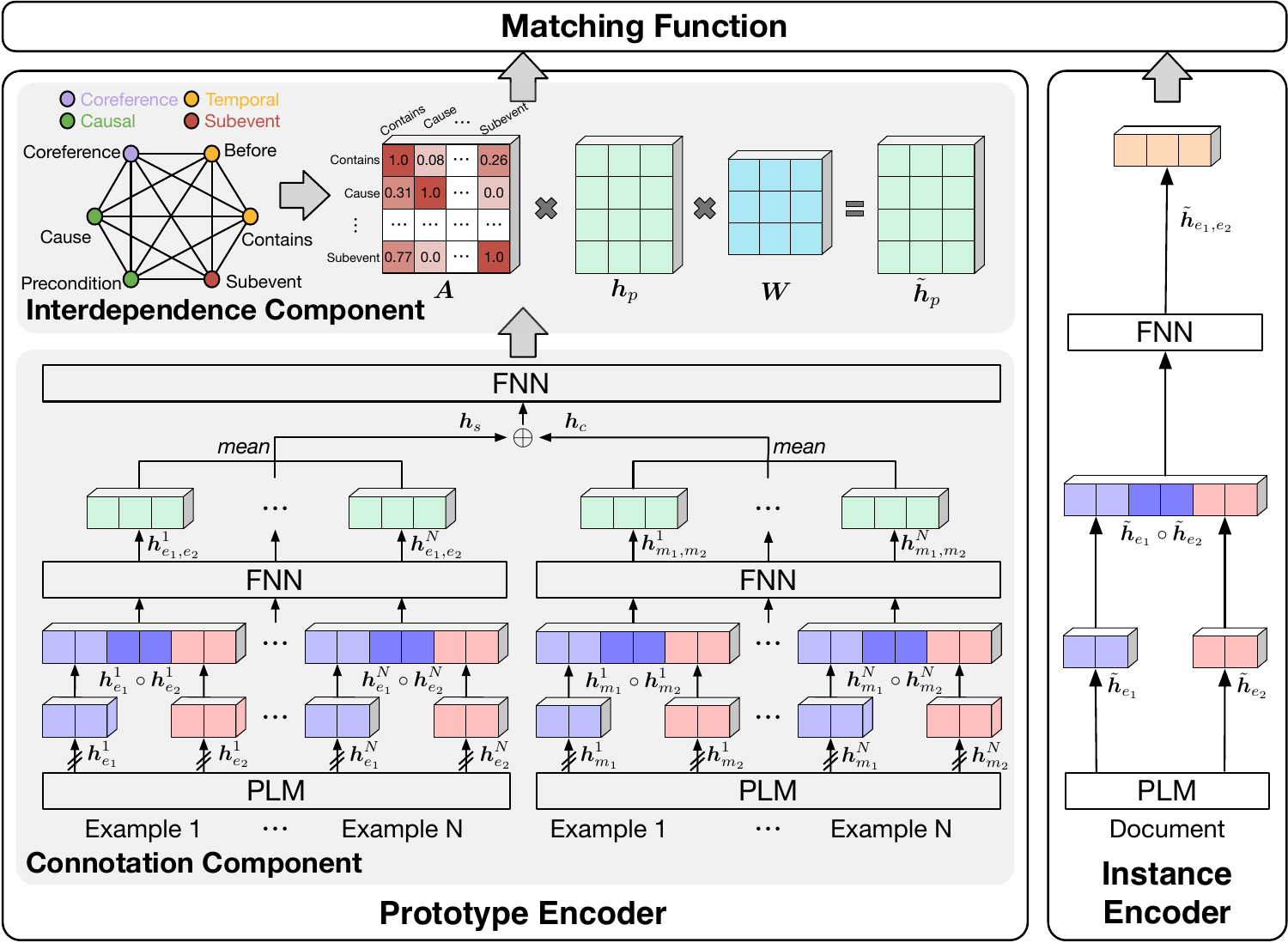}
  \caption{An illustration diagram of the proposed ProtoEM framework. ``//'' and $\oplus$ respectively denote the cessation of backpropagation and the concatenation operation.}
  \label{fig:model}
\end{figure*}

\section{Related Work}
\subsection{Event Relation Extraction}
Since the fundamental role of understanding event relations in natural language processing, extracting event relations has attracted extensive attention in the past few years.
Currently, there are four types of primary event relations receiving attention, namely temporal, causal, subevent, and coreference.
Early methods~\citep{Strotgen2010HeidelTimeHighQuality, Beamer2009UsingBigramEvent, Riaz2013BetterUnderstandingCausality} exploit lexical and syntactic features to improve performance.
Recently, most methods employ neural networks to encode event mentions and contexts.
Some methods~\citep{Cao2021KnowledgeEnrichedEventCausality, Liu2020KnowledgeEnhancedEvent, Hu2023SemanticStructureEnhanced} leverage the structural information or external knowledge to enrich the representations of events and capture the underlying connections between events.
Other methods~\citep{Ning2018JointReasoningTemporal, Chen2022ERGOEventRelational} explore properties such as symmetry and transitivity of event relations. 
In addition, several methods~\citep{Wang2020JointConstrainedLearning, Man2022SelectingOptimalContext} employ joint reasoning to model multiple types of event relations.
However, existing methods merely classify event relations into distinct categories, far from understanding the intrinsic semantics of relations.
Therefore, we attempt to utilize prototypes to represent event relations and explore their correlation.

\subsection{Prototype Learning}
The idea of prototypes could be traced to earlier studies in machine learning~\citep{Graf2009PrototypeClassificationInsightsb} and cognitive modeling and psychology~\citep{Reed1972PatternRecognitionCategorization, Rosch1976BasicObjectsNatural}.
Prototype learning assigns a representative prototype for each class, and class prediction is determined by measuring the similarity between an input and prototypes using a distance metric.
Prototype learning typically exhibits significant advantages in scenarios of few-shot learning and zero-shot learning~\citep{Snell2017PrototypicalNetworksFewshot, Pahde2021MultimodalPrototypicalNetworks, Huang2021BehaviorRegularizedPrototypical}.
For example, \citet{Gao2019HybridAttentionBasedPrototypical} propose hybrid attention-based prototypical networks for noisy few-shot relation classification.
\citet{Zhao2022KnowledgeEnhancedSelfSupervisedPrototypical} propose a knowledge-enhanced prototypical network for few-shot event detection.
In this work, we apply the idea of prototype learning to the ERE task within a supervised learning scenario.

\section{The ProtoEM Framework}
In this section, we introduce the proposed ProtoEM framework. 
The purpose of ProtoEM is to better comprehend event relations and match them with event pairs.
Thus, as shown in Figure~\ref{fig:model}, ProtoEM consists of three parts: a prototype encoder, an instance encoder, and a matching function.
The input to the prototype encoder consists of $N$ examples corresponding to each class of event relations.
This encoder employs the connotation component and interdependence component to aggregate two kinds of semantic information, i.e., the connotations of prototypes and the interdependence among prototypes, and later obtain prototype representations for each event relation.
The instance encoder takes the document as input, treats each event pair as an instance, and obtains their corresponding representations.
Finally, the matching function calculates the similarity score between instances and each prototype.

\subsection{Prototype Encoder}
The prototype encoder is specifically designed to obtain prototype representations for each event relation, and it consists of two components connected in series: the connotation component and the interdependence component.

\subsubsection{Connotation Component} 
For understanding event relations, the most intuitive semantic information comes from their typical and correctly applied examples.
We utilize prototypes to represent event relations and employ the connotation component to capture the connotations of prototypes.
Firstly, it selects $N$ examples corresponding to the current event relation and encodes these $N$ examples using PLM.
In each example, there is a piece of text containing two events, and there exists a corresponding event relation between them, such as the ``Cause'' relation.
After encoding the text using PLM, the representation of each event will be obtained by averaging the representations of all tokens corresponding to it.
The representations of both events will be fed into a Feedforward Neural Network (FNN), facilitating comprehensive interaction of event information.
Then, we obtain the representation $\boldsymbol{h}_{e_1,e_2}$ of the event pair:
\begin{equation}
\boldsymbol{h}_{e_1,e_2} = \sigma \left( \boldsymbol{W}_s [\boldsymbol{h}_{e_1} || \boldsymbol{h}_{e_2} || \boldsymbol{h}_{e_1} \circ \boldsymbol{h}_{e_2}] + \boldsymbol{b}_s\right),
\end{equation}
where $\boldsymbol{h}_{e_1}$ and $\boldsymbol{h}_{e_2}$ are the representations of events $e_1$ and $e_2$ in an event pair, respectively;
$\circ$ and $||$ denote the element-wise multiplication operation and the concatenation operation, respectively;
$W_s$ and $b_s$ are the weight matrix and bias, respectively;
$\sigma$ is the activation function (e.g., ReLU~\citep{Glorot2011DeepSparseRectifier}).

Drawing inspiration from prototype networks~\citep{Snell2017PrototypicalNetworksFewshot} and to minimize the impact of noisy information, we compute the average representation of event pairs for $N$ examples, thereby obtaining a holistic representation $\boldsymbol{h}_{s}$ of the examples:
\begin{equation}
 \boldsymbol{h}_{s} = \frac{1}{N} \sum_{i=1}^{N} \boldsymbol{h}_{e_1,e_2}^i ,
\end{equation}
where $\boldsymbol{h}_{e_1,e_2}^i$ is the representation of the event pair in the $i$-th example.

In addition to event information, we also leverage event-agnostic contextual patterns to provide additional information.
Specifically, a special token [mask] is employed to replace event tokens in the text, thereby excluding event-related information.
Similar to the previous process, we obtain event-agnostic contextual pattern information $\boldsymbol{h}_{c}$:
\begin{equation}
\boldsymbol{h}_{m_1,m_2} = \sigma \left( \boldsymbol{W}_s [\boldsymbol{h}_{m_1} || \boldsymbol{h}_{m_2} || \boldsymbol{h}_{m_1} \circ \boldsymbol{h}_{m_2}] + \boldsymbol{b}_s\right),
\end{equation}
\begin{equation}
 \boldsymbol{h}_{c} = \frac{1}{N} \sum_{i=1}^{N} \boldsymbol{h}_{m_1,m_2}^i ,
\end{equation}
where $\boldsymbol{h}_{m_1,m_2}^i$ is the representation of the event-agnostic context for the $i$-th example.

Subsequently, to aggregate the acquired representations of event pairs and contexts, we feed them into a FNN and obtain the prototype representations $\boldsymbol{h}_p$ for each event relation:
\begin{equation}
\boldsymbol{h}_{p} = \sigma \left( \boldsymbol{W_p} [\boldsymbol{h}_{s} || \boldsymbol{h}_{c}] + \boldsymbol{b}_p\right),
\end{equation}
where $\boldsymbol{W}_p$ and $\boldsymbol{b}_p$ are the weight matrix and bias, respectively.
It is worth noting that for the event relation "None" (indicating that this type of event relation does not exist between the two events), we do not utilize examples, as examples would introduce additional noise information.
We simply use the text 'None' for representation.

\subsubsection{Interdependence Component} 
Besides the connotations of prototypes, the interdependence among prototypes is also crucial for understanding the semantics of event relations.
To capture the dependence among prototypes, ProtoEM first constructs a dependency graph. 
Specifically, the dependency graph is a bidirectional complete graph with weighted edges, where nodes denote prototypes and edges represent the dependencies among prototypes.
To more accurately reflect the strength of dependencies among prototypes, we utilize the co-occurrence frequency of event relations within the training set as their dependency weights.
For example, in the interdependence component of Figure~\ref{fig:model}, the dependency strength between the prototypes ``\textit{Subevent}'' and ``\textit{Contains}'' is 0.77.
This means that among all event pairs with a ``\textit{Subevent}'' relation, 77\% of them also have a ``\textit{Contains}'' relation.
The specific dependency weight matrix is denoted as $\boldsymbol{A}$, and it will be shown in the appendix.
During the training process, the adjacency matrix is normalized, i.e., apart from the diagonal elements being 1, the sum of elements in each row is equal to 1, and the sum of elements in each column is also equal to 1.

GNNs have the capability to aggregate information from neighboring nodes to the specific node, making it well-suited for modeling interdependence among prototypes.
Therefore, we apply a Graph Convolutional Network (GCN) to aggregate
semantic information from neighbors of the prototypes.
Specifically, the message passing at layer $l\in[0,L-1]$ is conducted as follows:
\begin{equation}
    \boldsymbol{h}_{i}^{l+1} = \sigma \left( \sum_{j\in N_i} \boldsymbol{A}_{i,j} \boldsymbol{W}^{l} \boldsymbol{h}_{j}^{l} + \boldsymbol{W}_{0}^{l} \boldsymbol{h}_{i}^{l} \right),
\end{equation}
where $N_i$ denotes the set of neighbors of node $i$;
$\boldsymbol{h}_{i}^{l}$ and $\boldsymbol{h}_{j}^{l}$ denote the $l$-th layer representations of the nodes $i$ and $j$, respectively;
$\boldsymbol{A}_{i,j}$ represents the dependency weight between the prototype nodes $i$ and $j$;
$\boldsymbol{W}^{l}$ and $\boldsymbol{W}_{0}^{l}$ are weight matrices used to aggregate features of neighboring nodes and self-loop features in the $l$-th layer, respectively;
$\boldsymbol{h}_{i}^{0}$ and $\boldsymbol{h}_{j}^{0}$ are the initialized representations of the nodes which are obtained from the connotation component.

After capturing the interdependence among prototypes, the representations of prototypes are updated as $\tilde{\boldsymbol{h}}_{p}$ and then fed into the subsequent matching function.

\subsection{Instance Encoder}
ProtoEM adopts an instance encoder to encode the document and treats each event pair as an individual instance.
Specifically, the input of the instance encoder is a document $D = [x_t]_{t=1}^{L_D}$, where $x_t$ is the $t$-th token in the document $D$ and $L_D$ is the length of the document $D$. 
Then a PLM is employed to encode the whole document.
Due to the length limitation of the PLM in handling documents (cannot process documents exceeding 512 tokens), we adopt a sliding window mechanism to process the documents.
The document is divided into multiple spans, and if the current window is unable to accommodate a particular sentence, this sentence will be placed in the subsequent window.
The start and end positions of each window are marked with the special tokens [CLS] and [SEP], respectively.
As previously mentioned, after encoding through the PLM, the representation of each event will be obtained by averaging the representations of all tokens corresponding to it.
We then combine all events in pairs and feed each event pair into a multi-layer FNN.
The representation $\tilde{\boldsymbol{h}}_{e_1, e_2}$ of the event pair is obtained as follows:
\begin{equation}
\tilde{\boldsymbol{h}}_{e_1,e_2} = FNN(\tilde{\boldsymbol{h}}_{e_1} || \tilde{\boldsymbol{h}}_{e_2} || \tilde{\boldsymbol{h}}_{e_1} \circ \tilde{\boldsymbol{h}}_{e_2}),
\end{equation}
where $\tilde{\boldsymbol{h}}_{e_1}$ and $\tilde{\boldsymbol{h}}_{e_2}$ donate the representations of events $e_1$ and $e_2$ in an event pair, respectively.
Next, we feed the obtained representations of all event pairs into the subsequent matching function.

\subsection{Matching Function}
With the representations $\tilde{\boldsymbol{h}}_{p}$ of prototypes and the representations $\tilde{\boldsymbol{h}}_{e_1,e_2}$ of event pairs as input, the matching function calculates their matching similarity.
Same as the previous work~\citep{Liu2022PretrainingMatchUnified}, we adopt the Euclidean Distance to measure the similarity:
\begin{equation}
S(x, y_i) = - \left\| \tilde{\boldsymbol{h}}_{e_1,e_2} - \tilde{\boldsymbol{h}}_{p}^{i} \right\| ,
\end{equation}
where $\tilde{\boldsymbol{h}}_{q}^{i}$ is the representation of $i$-th prototype.
The corresponding probability for each relation is:
\begin{equation}
P(y_i | x;\theta) = \frac{\exp(S(x, y_i))}{\sum_{j=1}^{N} \exp(S(x, y_j))} ,
\end{equation}
where $N$ denotes the number of categories for each type of event relation.
Finally, the relation $\hat{y}$ with the maximal probability as the prediction result:
\begin{equation}
\hat{y} = \arg \max \limits_{y_i} P(y_i | x; \theta).
\end{equation}

\subsection{Parameter Learning}
For the four types of event relations, i.e., temporal, causal, subevent, and coreference, we adopt the cross-entropy loss as the loss function:
\begin{equation}
L(\theta)_k = - \sum_{i=1}^{N} \boldsymbol{I}(y_i)\log P(y_i | x;\theta),
\end{equation}
where $k$ denotes the $k$-th type of event relation;
$N$ indicates the number of categories for each type of event relation;
$\boldsymbol{I}(.)$ denotes the indicator function, $\boldsymbol{I}(y_i) $ equals 1 if $y_i$ is the golden class, otherwise $\boldsymbol{I}(y_i)$ equals 0.

Then, to facilitate knowledge sharing across multiple kinds of event relations and enhance the generalization ability of the model, following~\citet{Wang2022MAVENEREUnifiedLargescale}, we employ a joint training method to combine the losses of the four types of event relations.
The final loss of the model is represented as:
\begin{equation}
L(\theta) = \sum_{k} \lambda_{k} L(\theta)_k,
\end{equation}
where $\lambda_{k}$ indicates the weighting factor for the $k$-th loss.
The training objective is to minimize the above loss function.

\begin{table}[t]
    \centering
        \begin{tabular}{l | c c c}
        \toprule
        \textbf{MAVEN-ERE} & \textbf{Train} & \textbf{Valid} \\ 
        \midrule
        \# Document & 2,913 & 710 \\
        \# Mention & 73,939 & 17,780 \\
        \# TIMEX & 16,688 & 4,139 \\
        \# Temporal & 884,694 & 208,300 \\
        \# Causal & 53,358 & 13,624 \\
        \# Subevent & 20,101 & 6,099 \\
        \# Coreference & 30,934 & 8,058 \\
        \bottomrule
        \end{tabular}
    \caption{Dataset statistics of the MAVEN-ERE dataset on the event mention level. TIMEX denotes temporal expressions.
    }
    \label{table:dataset}
\end{table}

\begin{table*}[t]
    \centering
    \resizebox{\textwidth}{!}{
        \begin{tabular}{l | c c c c | c}
        \toprule
        \multirow{2}{*}{\textbf{Model}} & \textbf{Temporal} & \textbf{Causal} & \textbf{Subevent} & \textbf{Coreference} & \textbf{Overall} \\ \cmidrule{2-6}
        & \textbf{F1} & \textbf{F1} & \textbf{F1} & \textbf{F1} & \textbf{F1} \\
        \midrule
        SPEECH$^{\dag}$~\citep{Deng2023SPEECHStructuredPrediction} & 40.23 ± 0.34 & 16.31 ± 0.97 & 21.96 ± 1.24 & - & - \\
        ERGO~\citep{Chen2022ERGOEventRelational} & 51.46 ± 0.27 & 27.96 ± 0.65 & 23.39 ± 0.37 & 89.51 ± 0.08 & 48.08 ± 0.31 \\
        Joint$_\text{BERT}$ (Our implementation) & 51.12 ± 0.47 & 29.94 ± 0.46 & 22.98 ± 1.07 & 90.28 ± 0.10 & 48.58 ± 0.19 \\
        Joint$_\text{RoBERTa}$~\citep{Wang2022MAVENEREUnifiedLargescale} & 52.73 ± 0.14 & 30.81 ± 0.48 & 23.11 ± 0.65 & \textbf{90.80 ± 0.11} & 49.36 ± 0.27 \\
        \midrule
        \textbf{ProtoEM (ERGO)} & 51.26 ± 0.59 & 28.86 ± 0.36 & 25.35 ± 1.18 & 89.95 ± 0.11 & 48.85 ± 0.14 \\
        \textbf{ProtoEM (BERT)} & 52.85 ± 0.12 & 31.96 ± 0.24 & 29.73 ± 0.26 & 90.21 ± 0.05 & 51.19 ± 0.08 \\
        \textbf{ProtoEM (RoBERTa)} & \textbf{54.17 ± 0.13} & \textbf{33.93 ± 0.20} & \textbf{30.55 ± 0.33} & 90.20 ± 0.27 & \textbf{52.21 ± 0.02} \\
        \bottomrule
        \end{tabular}
    }
    \caption{Experimental results on the MAVEN-ERE validation set. $\dag$: evaluation only conducted on the validation set.
    }
    \label{table:dev}
\end{table*}

\begin{table*}[t]
    \centering
    \resizebox{\textwidth}{!}{
        \begin{tabular}{l | c c c c | c}
        \toprule
        \multirow{2}{*}{\textbf{Model}} & \textbf{Temporal} & \textbf{Causal} & \textbf{Subevent} & \textbf{Coreference} & \textbf{Overall} \\ \cmidrule{2-6}
        & \textbf{F1} & \textbf{F1} & \textbf{F1} & \textbf{F1} & \textbf{F1}\\
        \midrule
        GPT-4$^{\ddag}$~\citep{Yu2023KoLACarefullyBenchmarking} & 22.10 & 24.40 & 25.00 & 76.20 & 36.90 \\
        ERGO~\citep{Chen2022ERGOEventRelational} & 50.44 ± 2.31 & 22.51 ± 0.70 & 18.85 ± 0.86 & 90.44 ± 0.16 & 45.56 ± 0.45 \\
        Joint$_\text{BERT}$ (Our implementation) & 53.70 ± 0.61 & 30.43 ± 0.16 & 24.32 ± 0.84 & 91.32 ± 0.04 & 49.94 ± 0.22 \\
        Joint$_\text{RoBERTa}$~\citep{Wang2022MAVENEREUnifiedLargescale} & 56.00 ± 0.59 & 31.50 ± 0.42 & 27.50 ± 1.10 & \textbf{92.10} & 51.78\\
        \midrule
        \textbf{ProtoEM (ERGO)} & 50.71 ± 1.39 & 24.06 ± 1.18 & 21.28 ± 3.46 & 90.12 ± 0.05 & 46.54 ± 1.01 \\
        \textbf{ProtoEM (BERT)} & 55.50 ± 0.20 & 32.72 ± 0.31 & 32.04 ± 0.47 & 91.26 ± 0.23 & 52.88 ± 0.08 \\
        \textbf{ProtoEM (RoBERTa)} & \textbf{56.80 ± 1.00} & \textbf{34.55 ± 0.72} & \textbf{32.11 ± 0.74} & 91.60 ± 0.06 & \textbf{53.77 ± 0.29} \\
        \bottomrule
        \end{tabular}
    }
    \caption{Experimental results on the MAVEN-ERE test set. $\ddag$: evaluation conducted using a small subset of test instances (199 in total).
    }
    \label{table:test}
\end{table*}

\section{Experiments}
\subsection{Datasets}
We use the MAVEN-ERE~\citep{Wang2022MAVENEREUnifiedLargescale} dataset to evaluate the proposed framework.
MAVEN-ERE is the first unified large-scale dataset of event relations, annotated on English Wikipedia documents in the general domain.
This dataset is annotated with four types of event relations, including temporal, causal, subevent, and coreference relations.
Furthermore, the temporal relation is subdivided into ``Before'', ``Overlap'', ``Contains'', ``Simultaneous'', ``Ends-on'', and ``Begins-on'' relations.
The causal relation is subdivided into ``Precondition'' and ``Cause'' relations.
Subevent and coreference relations have no subdivision relations.
Each type of event relation also includes a ``None'' category.
The details of the dataset are presented in Table~\ref{table:dataset}.
Additionally, the test dataset is not publicly available, and we evaluate the model on an online platform\footnote{\href{https://codalab.lisn.upsaclay.fr/competitions/8691}{https://codalab.lisn.upsaclay.fr/competitions/8691}}.

\subsection{Evaluation Metrics}
For temporal, causal, and subevent relation extraction tasks, we adopt the standard micro-averaged precision, recall, and F1-score as evaluation metrics.
For event coreference resolution, following previous works~\citep{Choubey2017EventCoreferenceResolution, Lu2022EndtoendNeuralEvent}, we adopt MUC~\citep{Vilain1995ModelTheoreticCoreferenceScoring}, B$^3$~\citep{Bagga1998EntityBasedCrossDocumentCoreferencing}, CEAF$_e$~\citep{Luo2005CoreferenceResolutionPerformance} and BLANC~\citep{Recasens2011BlancImplementingRand} metrics. 
Due to space limitations, we report the average F1-score of all coreference metrics.
All results are the means and standard deviations of three independent experiments with distinct random seeds.

\subsection{Implementation Details}
In the experiments, the PLMs used in this paper are BERT-base~\citep{Devlin2019BERTPretrainingDeep} and RoBERTa-base~\citep{Liu2019RoBERTaRobustlyOptimized}.
The dimensions of prototypes and event pairs are set to 768, the same as the dimensions of tokens.
We select ReLU as the activation function with a dropout rate of 0.2.
The weighting factors $\lambda$ for the losses of coreference, temporal, causal, and subevent relations are set to 1.0, 2.0, 4.0, and 4.0, respectively.
The example size for each relation is set to 5; the number of GCN layers is set to 1.
We perform a grid search for the learning rates.
The learning rate for the PLM is set in \{1e-5, \textbf{2e-5}, 3e-5, 5e-5\}, while the learning rate for other parameters is set in \{1e-4, \textbf{3e-4}, 5e-4\}.
The AdamW~\citep{Loshchilov2019DecoupledWeightDecay} gradient strategy is used to optimize all parameters and the warm-up steps are set to 200.
The batch size is set to 8, and the model is trained for 50 epochs.
All the experiments are carried out on a 32GB Hygon DCU Z100 GPU.

\subsection{Baselines}
We apply the following representative methods as baselines: \textbf{SPEECH}~\citep{Deng2023SPEECHStructuredPrediction} is a method that models complex event structures with energy-based networks and efficiently represents events with event-centric hyperspheres;
\textbf{ERGO}~\citep{Chen2022ERGOEventRelational} is a method designed for Document-level Event Causality Identification and it captures the transitivity of event relations. We conduct the evaluation using the official implementation code and employ the RoBERTa as the PLM.
\textbf{Joint$_\text{RoBERTa}$}~\citep{Wang2022MAVENEREUnifiedLargescale} is the state-of-the-art (SOTA) method that performs joint training for multiple kinds of event relations, and we add a \textbf{Joint$_\text{BERT}$} model as baseline.
In addition, we also compare our approach with the \textbf{GPT-4}~\citep{OpenAI2023GPT4TechnicalReport} large language model that has attracted extensive attention.

\subsection{Main Experimental Results}
Tables~\ref{table:dev} and~\ref{table:test} present the experimental results on the validation and test sets of the MAVEN-ERE dataset.
Overall, our method outperforms all baselines in terms of the average F1-score across the four types of event relations.

Compared with the SOTA method, ProtoEM achieves 2.85\% and 1.99\% improvements in terms of average F1 scores on the validation and test sets, respectively.
Moreover, it achieves noticeable improvements across various event relations other than the coreference relation.
This can be attributed to our precise modeling of the prototypes corresponding to event relations, thereby achieving a more accurate semantic matching between event pairs and prototypes.
The experimental results demonstrate that thoroughly understanding event relations is highly beneficial for the ERE task. 

Compared with the SOTA method, ProtoEM achieves 7.44\% and 4.61\% improvements on the subevent relation of the validation and test sets, respectively.
One possible reason is that prototype-related information is more conducive to understanding complex event relations, as the subevent relation is the most challenging type to understand among the four types of event relations.
Utilizing some examples to comprehend the connotations of prototypes and to capture the interdependence among prototypes is essential for the comprehension of complex event relations.

In addition, ProtoEM achieves comparable results to the SOTA method in the coreference relation.
This is primarily because the coreference relation is bidirectional, while other relations are unidirectional.
To validate this hypothesis, we introduce an instance encoder to singly model the coreference relation, while temporal, causal, and subevent relations are modeled using another instance encoder.
The experimental results are presented in the second row of Table~\ref{table:framework}, and it shows that there is a significant improvement in coreference relation results when using an individual instance encoder to model the coreference relation.
Furthermore, the SOTA method employs a specially designed algorithm for the coreference resolution~\citep{Joshi2019BERTCoreferenceResolution}, and applying it to other event relations is challenging.


\begin{figure}[ht]
  \centering
  \includegraphics[width=\linewidth]{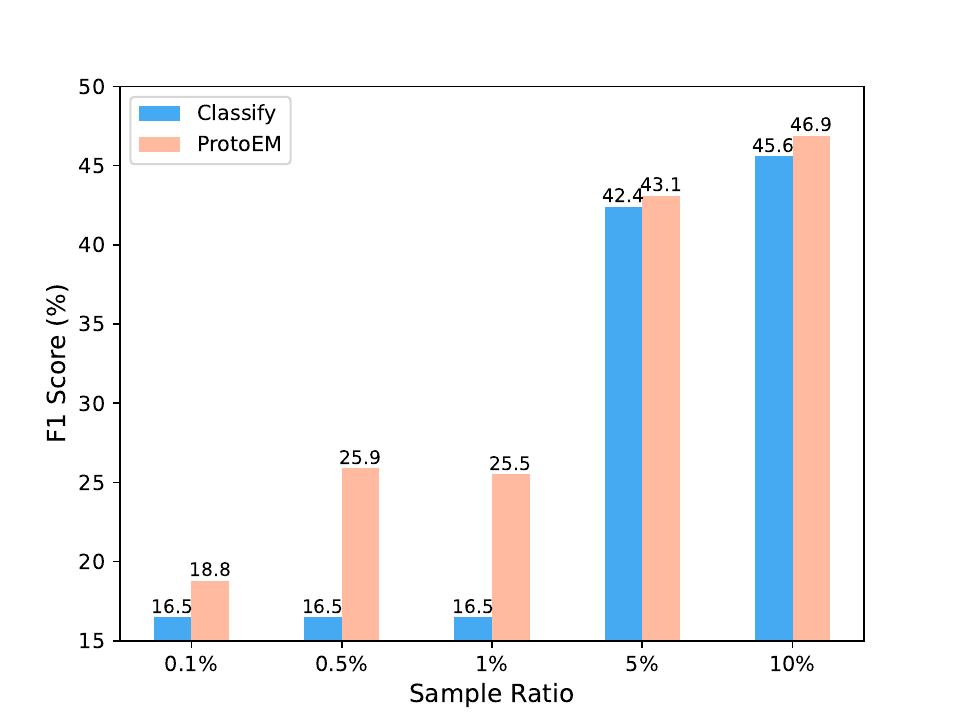}
  \caption{Experimental results under the low-resource setting.}
  \label{fig:low resource}
\end{figure}

\subsection{Experimental Results under the Low Resource Setting}

To illustrate the critical role of prototypes, we conduct experiments in the low-resource setting.
Specifically, we sample documents from the training set and evaluate the model on the entire validation set.
The experimental results are shown in Figure~\ref{fig:low resource}.
When there are very few training documents (0.5\%, approximately 15 documents), ProtoEM achieves a significant improvement of 9.4\% in average F1 score compared to the classification model.
This indicates that ProtoEM performs well under the low-resource setting.
The reason is that ProtoEM leverages examples to acquire more accurate representations of prototypes, allowing it to effectively extract event relations without requiring a large amount of training data.

\subsection{Ablation Study}
To illustrate the effect of two kinds of semantic information about prototypes, we conduct ablation experiments on the MAVEN-ERE validation set.
The results are presented in Table~\ref{table:ablation}.

\begin{table}[ht]
    \centering
    \resizebox{\linewidth}{!}{
        \begin{tabular}{l | c c c c | c}
        \toprule
        \multirow{2}{*}{\textbf{Model}} & \textbf{Temporal} & \textbf{Causal} & \textbf{Subevent} & \textbf{Coref.} & \textbf{Overall} \\ \cmidrule{2-6}
        & \textbf{F1} & \textbf{F1} & \textbf{F1} & \textbf{F1} & \textbf{F1} \\
        \midrule
        ProtoEM & 54.17 & 33.93 & 30.55 & 90.20 & 52.21 \\
        \quad -\emph{Graph}  & 53.87 & 33.75 & 27.60 & 90.16 & 51.35  \\
        \quad -\emph{Prototypes \& Graph}  & 53.03 & 31.62 & 24.96 & 90.40 & 50.00 \\
        \bottomrule
        \end{tabular}
    }
    \caption{Ablation results on the MAVEN-ERE validation set.
    -\emph{Graph} indicates ProtoEM conducts prototype matching without considering the interdependence among prototypes.
    -\emph{Prototypes \& Graph} indicates the model only utilizes the representations of event pairs encoded by the instance encoder for classification.
    }
    \label{table:ablation}
\end{table}

\subsubsection{Impact of the Interdependence of Prototypes}
To investigate the impact of the interdependence of prototypes on the results, we remove the interdependence component in the ProtoEM framework.
Compared with ProtoEM, -\emph{Graph} decreases by 0.86\% in terms of the average F1 score, with the subevent F1 score dropping by 2.95\%.
This shows that the interdependence among prototypes is beneficial for a more comprehensive understanding of these event relations. 
These dependencies are particularly valuable for comprehending complex event relations, such as the subevent relation.

\subsubsection{Impact of the Connotations of Prototypes}
Compared with -\emph{Prototypes \& Graph}, -\emph{Graph} achieves 1.35\% improvements in terms of the average F1 score.
This suggests that the connotations of prototypes are significant for understanding them and constitute a crucial component in enhancing the performance of the model.
Specifically, -\emph{Graph} achieves improvements of 0.84\%, 2.13\%, and 2.64\% on temporal, causal, and subevent relations, respectively.
This demonstrates the effectiveness of using examples to obtain prototype representations.

\subsection{Sub-module Analysis}
To further analyze the impact of each sub-module within the framework on the results, we conduct sub-module analysis experiments.
The results are presented in Table~\ref{table:different settings}.

\begin{table}[ht]
    \centering
    \resizebox{\linewidth}{!}{
        \begin{tabular}{l | c c c c | c}
        \toprule
        \multirow{2}{*}{\textbf{Model}} & \textbf{Temporal} & \textbf{Causal} & \textbf{Subevent} & \textbf{Coref.} & \textbf{Overall} \\ \cmidrule{2-6}
        & \textbf{F1} & \textbf{F1} & \textbf{F1} & \textbf{F1} & \textbf{F1} \\
        \midrule
        ProtoEM & 54.17 & 33.93 & 30.55 & 90.20 & 52.21 \\
        \quad Prototypes$_{random}$ & 53.84 & 32.83 & 26.09 & 90.54 & 50.83 \\
        \quad Prototypes$_{event}$ & 54.00 & 34.00 & 29.55 & 90.21 & 51.94 \\
        \quad Prototypes$_{context}$ & 54.37 & 33.94 & 29.39 & 90.31 & 52.00 \\
        \quad GCN$_{learned.W}$ & 53.85 & 34.14 & 27.87 & 90.04 & 51.48 \\
        \quad GCN$_{no.W}$ & 53.74 & 33.72 & 28.69 & 90.21 & 51.59 \\
        \bottomrule
        \end{tabular}
    }
    \caption{Experimental results on the MAVEN-ERE validation set under different settings.
    Prototypes$_{random}$ denotes the utilization of randomly initialized vectors as representations of prototypes.
    Prototypes$_{event}$ denotes only utilizes the event pair information in the connotation component.
    Prototypes$_{context}$ denotes only utilizes the event-agnostic context in the connotation component.
    GCN$_{learned.W}$ indicates that the weights within the dependency graph are learned by the model.
    GCN$_{no.W}$ indicates that all weights in the dependency graph are set to 1.
    }
    \label{table:different settings}
\end{table}

\subsubsection{Impact of Prototype Initialization}
Compared with ProtoEM, Prototypes$_{random}$ decreases by 1.38\% in terms of the average F1 score.
This once again demonstrates that utilizing the connotations of prototypes can result in a more precise understanding of event relations.
Furthermore, based on the results from Prototypes$_{event}$ and Prototypes$_{context}$, it can be observed that event pair information and event-agnostic contextual information are complementary.

\subsubsection{Impact of the Weights in the Dependency Graph}
Compared with ProtoEM, Graph$_{learned.W}$ decreases by 0.73\% in terms of the average F1 score.
This indicates that the model faces challenges in learning the interdependence among prototypes.
Compared with ProtoEM, Graph$_{no.W}$ shows a 0.62\% decrease in terms of the average F1 score.
This implies that there are differences in the strength of dependencies among prototypes, with some being closely related while others are unrelated.

\subsection{Detail Analysis}

\subsubsection{Analysis of the Architecture of the Framework}
To analyze the impact of the architecture of ProtoEM, we conduct experiments using different architectures.
The results are presented in Table~\ref{table:framework}.
Compared with ProtoEM, ``Two PLMs'' decreases by 1.24\% in terms of the average F1 score.
This might be attributed to the scarcity of examples within the prototype encoder, leading to inadequate training of the encoder.
Compared with ProtoEM, ``One PLM'' shows a 1.62\% decrease in terms of the average F1 score.
This could be because when using the same encoder to simultaneously encode prototypes and instances, prototypes can be affected by instances, making it difficult to learn an accurate prototype representation.

\begin{table}[t]
    \centering
    \resizebox{\linewidth}{!}{
        \begin{tabular}{l | c c c c | c}
        \toprule
        \multirow{2}{*}{\textbf{Model}} & \textbf{Temporal} & \textbf{Causal} & \textbf{Subevent} & \textbf{Coref.} & \textbf{Overall} \\ \cmidrule{2-6}
        & \textbf{F1} & \textbf{F1} & \textbf{F1} & \textbf{F1} & \textbf{F1} \\
        \midrule
        ProtoEM & 54.17 & 33.93 & 30.55 & 90.20 & 52.21 \\
        Coref. IND & 53.73 & 34.05 & 29.87 & 90.46 & 52.03 \\
        Two PLMs & 53.38 & 32.74 & 27.36 & 90.42 & 50.97 \\
        One PLM & 53.54 & 32.33 & 26.08 & 90.40 & 50.59 \\
        \bottomrule
        \end{tabular}
    }
    \caption{Experimental results of different architectures on the MAVEN-ERE validation set. 
    ``Coref. IND'' represents that the coreference relation is modeled using an individual instance encoder.
    ``Two PLMs'' denotes that the prototype encoder and instance encoder employ distinct PLMs that require parameter fine-tuning. 
    ``One PLM'' denotes that both encoders utilize an identical PLM that requires parameter fine-tuning.
    }
    \label{table:framework}
\end{table}

\subsubsection{Analysis of Example Selection Strategies}
\begin{figure}[ht]
  \centering
  \includegraphics[width=\linewidth]{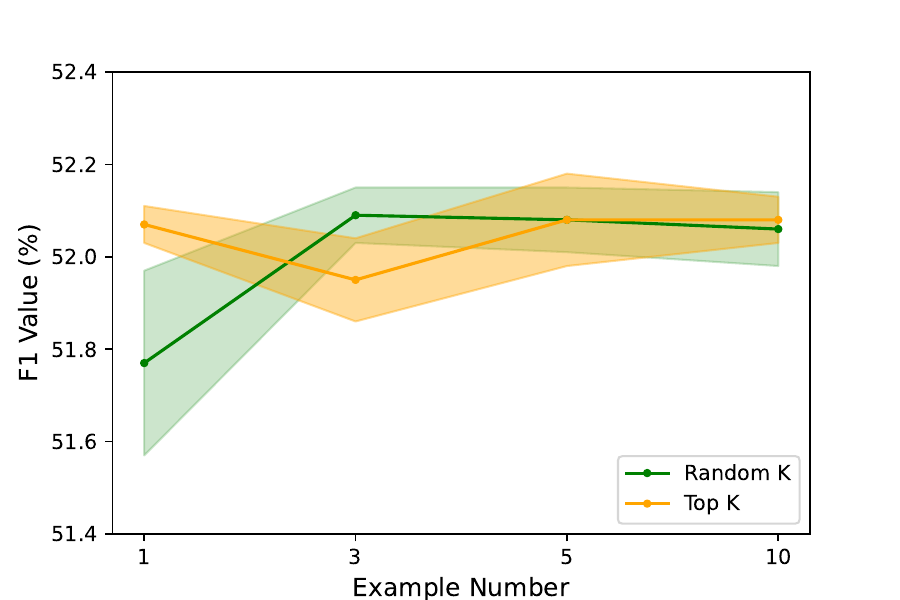}
  \caption{Experimental results under different example selection strategies.}
  \label{fig:example num}
\end{figure}

To explore the effect of example selection strategies on the model, two strategies are adopted for the experiment.
One strategy is fixed, selecting the top K event pairs with specific types of relations based on their frequency in the training data.
Another strategy involves randomness, randomly selecting K event pairs with specific types of relations from the training data.
The experimental results are shown in Figure~\ref{fig:example num}.
We can observe that when each prototype has only one example, the random selection strategy results in noticeable fluctuations in the results, while the top K selection strategy yields better and more stable results.
As the number of selected examples increases, both strategies exhibit similar mean and standard deviation values.
This suggests that the model possesses strong robustness and is relatively less affected by the example selection strategies.

\section{Conclusions}
In this paper, we proposed a Prototype-Enhanced Matching (ProtoEM) framework for ERE.
It utilizes the connotation component and the interdependence component to integrate two kinds of information, namely, the connotations of prototypes and the interdependence among prototypes.
Then it achieves ERE through a matching process between prototypes and event pairs.
Experimental results on the MAVEN-ERE dataset demonstrate that the proposed ProtoEM framework can effectively represent the prototypes of event relations, which is beneficial for the ERE task.

\section{Bibliographical References}\label{sec:reference}

\bibliographystyle{lrec-coling2024-natbib}
\bibliography{lrec-coling2024-example}



\appendix

\section{Supplementary Model Details}
The weight matrix among prototypes is displayed in Table~\ref{table:weight}.

\section{Detailed Experimental Results}
Tables~\ref{table:dev relation} and~\ref{table:test relation} present the results of temporal, causal, and subevent relations on the validation and test sets, respectively.
Tables~\ref{table:dev coref} and~\ref{table:test coref} present the results of the coreference relation on the validation and test sets, respectively.

\begin{table*}[ht]
    \centering
    \resizebox{\textwidth}{!}{
        \begin{tabular}{l | c c c c c c c c c c c c c c}
        \toprule
         Training set  &  None(coref.)  &  Coref  &  None(temporal)  &  Before  &  Overlap  &  Contains  &  Simultaneous  &  Ends-on  &  Begins-on  &  None(causal)  &  Precondition  &  Cause  &  None(subevent)  &  Subevent  \\ 
        \midrule
         None(coref.)  & 1.00 &	0.00 &	0.78 &	0.19 &	0.00 &	0.02 &	0.00 &	0.00 &	0.00 &	0.98 	& 0.02 &	0.01 &	0.99 &	0.01   \\
         Coref  & 0.00 & 	1.00 & 	1.00 & 	0.00 	& 0.00 & 	0.00 & 	0.00 	& 0.00 	& 0.00 & 	1.00 & 	0.00 	& 0.00 & 	1.00 & 	0.00   \\
         None(temporal)  & 0.98& 	0.02 &	1.00 &	0.00 &	0.00 &	0.00 &	0.00 &	0.00 &	0.00 &	1.00 &	0.00 	&0.00 &	1.00 &	0.00   \\
         Before  & 1.00  &	0.00  &	0.00  &	1.00  &	0.00  &	0.00  &	0.00  &	0.00  &	0.00  &	0.92 	 &0.06  &0.01  &	1.00  &	0.00   \\
         Overlap  & 1.00 &	0.00 &	0.00 &0.00 &	1.00 &	0.00 &	0.00 &	0.00 	&0.00 &	0.77 &	0.13 &	0.09 &	1.00 &	0.00   \\
         Contains  & 1.00 &	0.00 	&0.00 &	0.00 &	0.00 &	1.00 &	0.00 &	0.00 &	0.00 &	0.87 &	0.05 &	0.08 &	0.74 &	0.26   \\
         Simultaneous  & 1.00 &	0.00 &	0.00 	&0.00& 	0.00 &	0.00 &	1.00 &	0.00 &	0.00 &	1.00 &	0.00 &	0.00 &	1.00 	&0.00   \\
         Ends-on  & 1.00 &	0.00 &	0.00 &	0.00 &	0.00 	&0.00 &	0.00 &	1.00 	&0.00 &	0.85 &	0.15 &	0.00 &	1.00 	& 0.00   \\
         Begins-on  & 1.00 &	0.00 &	0.00 &	0.00 	&0.00 &	0.00 &	0.00 	&0.00 	&1.00 &	0.98 &	0.01 &	0.00 &	0.87 	&0.13   \\
         None(causal)  & 0.99 &	0.01 &	0.79 &	0.18 &	0.00 &	0.02 &	0.00 	&0.00 &	0.00 &1.00 &	0.00 &	0.00 	&0.99 &	0.01   \\
         Precondition  & 1.00 &	0.00 &	0.12 &	0.79 &	0.01 &	0.08 &	0.00 &	0.00 &	0.00 	&0.00 &	1.00 &	0.00 	&1.00 &	0.00   \\
         Cause  & 1.00 &	0.00 &	0.18 &	0.49 &	0.02 &	0.31 &	0.00 &	0.00 &	0.00 &	0.00 &	0.00 &	1.00 &	1.00 &	0.00   \\
         None(subevent)  & 0.99  &	0.01  &	0.79  &	0.19  &	0.00  &	0.02  &	0.00 	 &0.00  &	0.00  &	0.98 	 &0.02 	 &0.01 	 &1.00 	 &0.00   \\
         Subevent  & 1.00 &	0.00 &	0.22 &	0.00 &	0.00 &	0.77 &	0.00 &	0.00 &	0.00 &	1.00 &	0.00 &	0.00 &	0.00 	&1.00   \\

        \bottomrule
        
        \end{tabular}
    }
    \caption{The weight matrix among prototypes.
    }
    \label{table:weight}
\end{table*}

\begin{table*}[ht]
    \centering
    \resizebox{\textwidth}{!}{
        \begin{tabular}{l | c c c | c c c | c c c}
        \toprule
        \multirow{2}{*}{\textbf{Model}} & \multicolumn{3}{|c|}{\textbf{Temporal}} & \multicolumn{3}{|c|}{\textbf{Causal}} & \multicolumn{3}{|c}{\textbf{Subevent}} \\ \cmidrule{2-10}
        & \textbf{P} & \textbf{R} & \textbf{F1} & \textbf{P} & \textbf{R} & \textbf{F1} & \textbf{P} & \textbf{R} & \textbf{F1} \\
        \midrule
        ERGO & 49.99$_{1.49}$ & 53.11$_{2.13}$ & 51.46$_{0.27}$ & 34.13$_{1.54}$ & 23.76$_{1.66}$ & 27.96$_{0.65}$ & 31.20$_{0.66}$ & 18.72$_{0.66}$ & 23.39$_{0.37}$ \\
        Joint$_\text{BERT}$ & 50.46$_{3.61}$ & 52.22$_{4.35}$ & 51.12$_{0.47}$ & 31.68$_{0.92}$ & 28.46$_{1.62}$ & 29.94$_{0.46}$ & 30.34$_{0.34}$ & 18.51$_{1.36}$ & 22.98$_{1.07}$ \\
        Joint$_\text{RoBERTa}$ & 50.65$_{0.73}$ & 55.01$_{0.83}$ & 52.73$_{0.14}$ & 32.52$_{1.29}$ & 29.34$_{1.57}$ & 30.81$_{0.48}$ & 28.24$_{0.25}$ & 19.56$_{0.82}$ & 23.11$_{0.65}$ \\
        \midrule
        \textbf{ProtoEM (ERGO)} & 49.74$_{0.79}$ & 52.93$_{2.10}$ & 51.26$_{0.59}$ & 32.45$_{1.83}$ & 26.03$_{0.64}$ & 28.86$_{0.36}$ & 25.81$_{4.06}$ & 26.29$_{6.82}$ & 25.35$_{1.18}$ \\
        \textbf{ProtoEM (BERT)} & 47.58$_{0.43}$ & 59.43$_{0.55}$ & 52.85$_{0.12}$ & 28.34$_{0.18}$ & 36.65$_{0.33}$ & 31.96$_{0.24}$ & 22.93$_{1.32}$ & 42.66$_{3.85}$ & 29.73$_{0.26}$ \\
        \textbf{ProtoEM (RoBERTa)} & 50.21$_{1.02}$ & 58.85$_{1.37}$ & 54.17$_{0.13}$ & 28.59$_{0.55}$ & 41.77$_{1.66}$ & 33.93$_{0.20}$ & 25.21$_{2.96}$ & 40.64$_{9.24}$ & 30.55$_{0.33}$ \\
        \bottomrule
        
        \end{tabular}
    }
    \caption{Experimental results on the MAVEN-ERE validation set for temporal, causal, and subevent relations.
    }
    \label{table:dev relation}
\end{table*}

\begin{table*}[ht]
    \centering
    \resizebox{\textwidth}{!}{
        \begin{tabular}{l | c c c | c c c | c c c}
        \toprule
        \multirow{2}{*}{\textbf{Model}} & \multicolumn{3}{|c|}{\textbf{Temporal}} & \multicolumn{3}{|c|}{\textbf{Causal}} & \multicolumn{3}{|c}{\textbf{Subevent}} \\ \cmidrule{2-10}
        & \textbf{P} & \textbf{R} & \textbf{F1} & \textbf{P} & \textbf{R} & \textbf{F1} & \textbf{P} & \textbf{R} & \textbf{F1} \\
        \midrule
        ERGO & 57.61$_{2.77}$ & 45.23$_{5.48}$ & 50.44$_{2.31}$ & 38.27$_{2.11}$ & 15.99$_{1.07}$ & 22.51$_{0.70}$ & 35.73$_{2.98}$ & 12.86$_{1.01}$ & 18.85$_{0.86}$ \\
        Joint$_\text{BERT}$ & 54.58$_{3.54}$ & 53.20$_{4.23}$ & 53.70$_{0.61}$ & 33.60$_{0.29}$ & 27.81$_{0.46}$ & 30.43$_{0.16}$ & 32.45$_{2.26}$ & 19.49$_{0.75}$ & 24.32$_{0.84}$ \\
        Joint$_\text{RoBERTa}$ & 55.40$_{0.91}$ & 56.60$_{1.52}$ & 56.00$_{0.59}$ & 33.80$_{1.00}$ & 29.50$_{0.83}$ & 31.50$_{0.42}$ & 29.80$_{1.76}$ & 25.60$_{1.57}$ & 27.50$_{1.10}$ \\
        \midrule
        \textbf{ProtoEM (ERGO)} & 55.22$_{1.98}$ & 47.06$_{3.65}$ & 50.71$_{1.39}$ & 37.00$_{2.84}$ & 17.95$_{1.89}$ & 24.06$_{1.18}$ & 29.79$_{2.75}$ & 17.25$_{5.66}$ & 21.28$_{3.46}$ \\
        \textbf{ProtoEM (BERT)} & 52.82$_{0.84}$ & 58.51$_{1.24}$ & 55.50$_{0.20}$ & 26.90$_{1.26}$ & 41.90$_{2.34}$ & 32.72$_{0.31}$ & 25.63$_{1.21}$ & 43.14$_{4.40}$ & 32.04$_{0.47}$ \\
        \textbf{ProtoEM (RoBERTa)} & 55.32$_{1.63}$ & 58.53$_{3.68}$ & 56.80$_{1.00}$ & 31.11$_{0.87}$ & 38.94$_{2.54}$ & 34.55$_{0.72}$ & 29.47$_{4.58}$ & 36.46$_{5.33}$ & 32.11$_{0.74}$ \\
        \bottomrule
        
        \end{tabular}
    }
    \caption{Experimental results on the MAVEN-ERE test set for temporal, causal, and subevent relations.
    }
    \label{table:test relation}
\end{table*}

\begin{table*}[ht]
    \centering
    \resizebox{\textwidth}{!}{
        \begin{tabular}{l | c c c | c c c | c c c | c c c}
        \toprule
        \multirow{2}{*}{\textbf{Model}} & \multicolumn{3}{|c|}{\textbf{MUC}} & \multicolumn{3}{|c|}{\textbf{B$^3$}} & \multicolumn{3}{|c|}{\textbf{CEAF$_e$}} & \multicolumn{3}{|c}{\textbf{BLANC}} \\ \cmidrule{2-13}
        & \textbf{P} & \textbf{R} & \textbf{F1} & \textbf{P} & \textbf{R} & \textbf{F1} & \textbf{P} & \textbf{R} & \textbf{F1} & \textbf{P} & \textbf{R} & \textbf{F1}\\
        \midrule
        ERGO & 78.53$_{1.75}$ & 75.32$_{2.20}$ & 76.86$_{0.36}$ & 97.52$_{0.31}$ & 97.67$_{0.24}$ & 97.59$_{0.03}$ & 97.11$_{0.22}$ & 97.47$_{0.21}$ & 97.29$_{0.01}$ & 83.29$_{0.93}$ & 90.03$_{1.36}$ & 86.32$_{0.03}$ \\
        Joint$_\text{BERT}$ & 79.92$_{1.49}$ & 77.48$_{1.93}$ & 78.66$_{0.37}$ & 97.65$_{0.28}$ & 97.81$_{0.17}$ & 97.73$_{0.06}$ & 97.34$_{0.18}$ & 97.60$_{0.19}$ & 97.47$_{0.02}$ & 84.61$_{1.22}$ & 90.43$_{0.92}$ & 87.27$_{0.34}$ \\
        Joint$_\text{RoBERTa}$ & 80.12$_{0.12}$ & 79.29$_{0.85}$ & 79.70$_{0.37}$ & 97.78$_{0.07}$ & 97.89$_{0.08}$ & 97.83$_{0.01}$ & 97.52$_{0.08}$ & 97.61$_{0.02}$ & 97.57$_{0.03}$ & 86.64$_{0.77}$ & 89.74$_{0.69}$ & 88.11$_{0.21}$ \\
        \midrule
        \textbf{ProtoEM (ERGO)} & 79.26$_{0.68}$ & 76.54$_{0.58}$ & 77.87$_{0.29}$ & 97.57$_{0.07}$ & 97.79$_{0.04}$ & 97.68$_{0.03}$ & 97.24$_{0.07}$ & 97.54$_{0.08}$ & 97.39$_{0.04}$ & 83.70$_{0.32}$ & 90.72$_{0.25}$ & 86.87$_{0.14}$ \\
        \textbf{ProtoEM (BERT)} & 77.26$_{0.83}$ & 80.44$_{1.18}$ & 78.81$_{0.22}$ & 97.19$_{0.20}$ & 98.10$_{0.16}$ & 97.64$_{0.03}$ & 97.59$_{0.12}$ & 97.22$_{0.11}$ & 97.41$_{0.01}$ & 83.12$_{0.73}$ & 91.90$_{0.95}$ & 86.97$_{0.08}$ \\
        \textbf{ProtoEM (RoBERTa)} & 77.79$_{0.72}$ & 79.83$_{1.68}$ & 78.79$_{0.71}$ & 97.30$_{0.16}$ & 98.06$_{0.18}$ & 97.68$_{0.04}$ & 97.51$_{0.19}$ & 97.27$_{0.10}$ & 97.39$_{0.09}$ & 83.23$_{0.43}$ & 91.67$_{1.22}$ & 86.95$_{0.26}$ \\
        \bottomrule
        
        \end{tabular}
    }
    \caption{Experimental results on the MAVEN-ERE validation set for the coreference relation.
    }
    \label{table:dev coref}
\end{table*}

\begin{table*}[ht]
    \centering
    \resizebox{\textwidth}{!}{
        \begin{tabular}{l | c c c | c c c | c c c | c c c}
        \toprule
        \multirow{2}{*}{\textbf{Model}} & \multicolumn{3}{|c|}{\textbf{MUC}} & \multicolumn{3}{|c|}{\textbf{B$^3$}} & \multicolumn{3}{|c|}{\textbf{CEAF$_e$}} & \multicolumn{3}{|c}{\textbf{BLANC}} \\ \cmidrule{2-13}
        & \textbf{P} & \textbf{R} & \textbf{F1} & \textbf{P} & \textbf{R} & \textbf{F1} & \textbf{P} & \textbf{R} & \textbf{F1} & \textbf{P} & \textbf{R} & \textbf{F1}\\
        \midrule
        ERGO & 81.04$_{2.87}$ & 75.74$_{2.82}$ & 78.23$_{0.32}$ & 97.98$_{0.45}$ & 97.82$_{0.29}$ & 97.90$_{0.09}$ & 97.29$_{0.26}$ & 97.83$_{0.32}$ & 97.56$_{0.04}$ & 86.36$_{2.08}$ & 90.09$_{1.77}$ & 88.07$_{0.41}$ \\
        Joint$_\text{BERT}$ & 81.72$_{1.58}$ & 79.40$_{1.59}$ & 80.52$_{0.08}$ & 97.96$_{0.27}$ & 98.11$_{0.18}$ & 98.03$_{0.04}$ & 97.68$_{0.16}$ & 97.92$_{0.17}$ & 97.80$_{0.01}$ & 86.40$_{1.20}$ & 91.95$_{1.39}$ & 88.94$_{0.12}$ \\
        Joint$_\text{RoBERTa}$ & 81.40$_{1.64}$ & 82.80$_{1.56}$ & 82.10$_{0.43}$ & 98.00$_{0.27}$ & 98.30$_{0.18}$ & 98.20$_{0.11}$ & 98.00$_{0.13}$ & 97.80$_{0.21}$ & 97.90$_{0.09}$ & 88.80$_{1.05}$ & 91.40$_{1.15}$ & 90.20$_{0.27}$ \\
        \midrule
        \textbf{ProtoEM (ERGO)} & 81.05$_{1.84}$ & 74.23$_{2.16}$ & 77.46$_{0.35}$ & 98.01$_{0.29}$ & 97.72$_{0.18}$ & 97.86$_{0.05}$ & 97.14$_{0.21}$ & 97.85$_{0.19}$ & 97.49$_{0.01}$ & 85.96$_{1.18}$ & 89.60$_{1.01}$ & 87.66$_{0.19}$ \\
        \textbf{ProtoEM (BERT)} & 78.37$_{1.67}$ & 82.88$_{0.89}$ & 80.55$_{0.47}$ & 97.47$_{0.21}$ & 98.41$_{0.08}$ & 97.94$_{0.07}$ & 97.93$_{0.03}$ & 97.44$_{0.26}$ & 97.68$_{0.11}$ & 85.19$_{0.59}$ & 93.46$_{0.51}$ & 88.87$_{0.40}$ \\
        \textbf{ProtoEM (RoBERTa)} & 79.86$_{1.74}$ & 82.70$_{2.13}$ & 81.22$_{0.10}$ & 97.71$_{0.29}$ & 98.39$_{0.19}$ & 98.05$_{0.06}$ & 97.91$_{0.17}$ & 97.60$_{0.25}$ & 97.75$_{0.04}$ & 86.37$_{0.86}$ & 92.96$_{0.66}$ & 89.37$_{0.24}$ \\
        \bottomrule
        
        \end{tabular}
    }
    \caption{Experimental results on the MAVEN-ERE test set for the coreference relation.
    }
    \label{table:test coref}
\end{table*}

\end{document}